\documentclass{clv2}

\issue{1}{1}{2006}

\usepackage{graphicx}
\usepackage{algorithm}
\usepackage{algorithmic}
\usepackage{mathtools}
\usepackage{hyperref}
\usepackage{mdframed}
\usepackage{courier}

\runningtitle{Boost Phrase-level with Review-level Sentiment Analysis}

\runningauthor{Yongfeng Zhang et al.}

\begin{document}

\title{Boost Phrase-level Polarity Labelling with Review-level Sentiment Classification\thanks{Part of this work was supported by Natural Science Foundation (60903107, 61073071) and National High Technology Research and Development (863) Program (2011AA01A205) of China, and the third author is sponsored by the National Science Foundation (IIS-0713111). The opinions, findings, suggestions or conclusions expressed in this paper are the authors', and do not necessarily reflect those of the sponsors.}\thanks{This paper is an extended version of the work \textit{Do users rate or review? boost phrase-level sentiment labeling with review-level sentiment classification} in SIGIR'14.}}

\author{Yongfeng Zhang}
\affil{Department of Computer Science\\ Tsinghua University\\ yongfeng14@mails.tsinghua.edu.cn}

\author{Min Zhang}
\affil{Department of Computer Science\\ Tsinghua University\\ z-m@tsinghua.edu.cn}

\author{Yiqun Liu}
\affil{Department of Computer Science\\ Tsinghua University\\ yiqunliu@tsinghua.edu.cn}

\author{Shaoping Ma}
\affil{Department of Computer Science\\ Tsinghua University\\ msp@tsinghua.edu.cn}

\maketitle

\begin{abstract}
Sentiment analysis on user reviews helps to keep track of user reactions towards products, and make advices to users about what to buy. 
State-of-the-art review-level sentiment classification techniques could give pretty good precisions of above $90\%$. 
However, current phrase-level sentiment analysis approaches might only give sentiment polarity labelling precisions of around $70\%\sim80\%$, which is far from satisfaction and restricts its application in many practical tasks.

In this paper, we focus on the problem of phrase-level sentiment polarity labelling and attempt to bridge the gap between phrase-level and review-level sentiment analysis. We investigate the inconsistency between the numerical star ratings and the sentiment orientation of textual user reviews. Although they have long been treated as identical, which serves as a basic assumption in previous work, we find that this assumption is not necessarily true. 

We further propose to leverage the results of review-level sentiment classification to boost the performance of phrase-level polarity labelling using a novel constrained convex optimization framework. Besides, the framework is capable of integrating various kinds of information sources and heuristics, while giving the global optimal solution due to its convexity. Experimental results on both English and Chinese reviews show that our framework achieves high labelling precisions of up to $89\%$, which is a significant improvement from current approaches.
\end{abstract}

\section{Introduction}
Sentiment analysis techniques could be classified into three levels according to the different granularities on which the analysis is conducted, i.e., document-level, sentence-level and phrase-level \cite{survey1}. When analyzing user reviews, document-level sentiment analysis is also referred to as review-level sentiment analysis. Figure \ref{fig:amazon-review} shows a user review on an Apple iPhone product, which is extracted from Amazon.com.

Review-level and sentence-level sentiment analysis attempt to determine the overall sentiment orientation of a review or sentence. In phrase-level sentiment analysis, however, we are particularly interested in those phrases that describe some features or aspects of products, which, in Figure \ref{fig:amazon-review} for example, are the phrases \textit{service} and \textit{phone quality}. The user used \textit{excellent} to modify the product feature \textit{service}, and \textit{perfect} for \textit{phone quality}. Both pairs express user's positive sentiment on the corresponding feature. In this work, we use Feature word (F) to represent the words or phrases that describe specific product features, use Opinion word (O) for the words or phrases expressing users' sentiments towards feature words, and use Sentiment polarity (S) for the sentiment orientation of a Feature-Opinion (FO) pair.

\begin{figure}[t]
\begin{mdframed}[innermargin=0.5cm,outermargin=0.5cm]\small
Star Rating: $\star\star\star\star\star$\\
Review: \textit{I am very happy to have bought this phone from Amazon and the \textbf{service} rendered from the seller is \textbf{excellent}. \textbf{Phone quality} is \textbf{perfect} as new though I bought a used one. Care to their customers is something a key strategy the seller has followed. I would like to deal again with the same group in the near future and recommend to others highly. Thank you.}
\end{mdframed}
\caption{A sample user review towards an Apple iPhone 5 product from Amazon.com, which consists of a numerical star rating (5 stars here) and a piece of review text. The feature-opinion word pairs (service, excellent) and (phone quality, perfect) could be extracted from the review text to represent user attitudes (opinion words) towards specific product features (feature words).}
\label{fig:amazon-review}
\vspace{-10pt}
\end{figure}

The construction of a sentiment lexicon is of key importance in phrase-level sentiment analysis \cite{survey-lexicon,lexicon1,lexicon2,optimization}. Each entry in the lexicon is an FO pair together with the corresponding sentiment polarity, represented by (F,O,S) \cite{extract,optimization,survey-lexicon}. For example, the entries (\textit{service, excellent, positive}) and (\textit{phone quality, perfect, positive}) could be extracted in the review of Figure \ref{fig:amazon-review}. The underlying reason for such approaches is the observation that the sentiment polarity of opinion words could be \textit{contextual} \cite{contextual,optimization}, which means that the same opinion word could lead to different sentiment orientations when used to modify different feature words. For example, the opinion word \textit{high} has a positive sentiment when modifying the feature word \textit{quality}, yet has a negative sentiment when accompanied by \textit{noise}.

%

However, current phrase-level sentiment lexicon construction approaches may only give sentiment polarity labelling (determining the S for an FO pair) precisions of around $70\%\sim80\%$ \cite{survey1,contextual,lexicon1,lexicon2}. Although the literature has shown that the overall sentiment classification precisions of a whole sentence or review could be reasonably high even the sentiment lexicon is not that accurate \cite{compare2,turney2002,lexicon1,lexicon2}, we argue that the problem of constructing an accurate sentiment lexicon itself is important, because the use of a sentiment lexicon might not be limited to aggregating the overall sentiment of sentences or reviews. In fact, it can be used in many promising tasks, such as word of mouth tracking of brands, tools for product design and optimization, and feature-level product search or recommendation.

Previous work on phrase-level sentiment analysis and lexicon construction \cite{optimization,bayesian,peanut} assumes that the accompanied user rating indicates the overall sentiment orientation of a review text. However, we would like to point out according to our experiments on user rating analysis that, the star ratings might not be a kind of reliable signal in this task, and a substantial amount of users tend to make similar or even the same ratings continuously, regardless of the review text that they comment on a specific product.

In this paper, however, we propose to boost the process of phrase-level sentiment polarity labelling in a reverse way, which is to use review-level sentiment classification results as a heuristic for phrase-level polarity labelling. State-of-the-art review-level sentiment classification techniques, even the unsupervised approaches, can give pretty good precisions of above $90\%$ \cite{survey1,compare1,chinese1,selfsuper,chinese2}, which could be reliable to help boost the performance of phrase-level sentiment polarity labelling. Here, we mainly focus our attention on unsupervised review-level sentiment classification techniques because of the fact that they need no manually annotated training data, which makes them domain-independent. Besides, they can achieve comparable or even better performance compared with supervised approaches, especially in some scenarios such as online product reviews \cite{survey1,compare1}.

We design a two-stage process for phrase-level polarity labelling. In the first stage, the overall sentiment orientations of the product reviews in the corpus are labeled using a review-level sentiment classifier. In the second stage, we extract feature-opinion pairs from the corpus, then use the overall sentiment orientations of reviews as constraints to learn the sentiment polarities of these pairs automatically using a novel optimization framework. Experimental results on both English and Chinese review datasets show that our framework improves the precision of phrase-level sentiment polarity labelling significantly, which means that it might be promising to leverage sentence- or review-level sentiment analysis techniques to boost the performance of phrase-level sentiment analysis tasks. The main contributions of this paper are as follows:

\begin{itemize}
\setlength{\topsep}{0ex}
\setlength{\itemsep}{0ex}
\setlength{\parsep}{0ex}
\setlength{\itemsep}{0ex}
	\item We investigate the phenomenon of inconsistency between the numerical star ratings and the sentiment polarities of textual user reviews though extensive experimental studies.
	\item We propose to leverage review-level sentiment analysis techniques to boost the performance of phrase-level sentiment polarity labelling in sentiment lexicon construction tasks.
	\item We formally define the problem of phrase-level sentiment polarity labelling as a constrained convex optimization problem and design iterative optimization algorithms for model learning, where the global optimal solution is guaranteed.
	\item Through a comprehensive experimental study on both English and Chinese datasets, the effectiveness of the proposed framework is verified.
\end{itemize}

The remainder of this paper will be structured as follows. Section \ref{sec:relate} reviews some related work, and Section \ref{sec:define} formally defines the problem that we investigate. In Section \ref{sec:framework}, we propose our phrase-level sentiment polarity labelling framework, followed by the experimental results in Section \ref{sec:experiment}. We conclude the work  in Section \ref{sec:conclusions}.


\section{Related work}\label{sec:relate}
With the rapid growth of e-commerce, social networks and online discussion forums, the web has been rich in user-generated free-text data, where users express various attitudes towards products or events, which have been attracting researchers into Sentiment Analysis \cite{survey1,survey2}. Sentiment analysis plays an important role in many applications, including opinion retrieval \cite{om}, word-of-mouth tracking \cite{wom}, and opinion oriented document summarization \cite{os1,hu-kdd04}, etc.

One of the core tasks in sentiment analysis is to determine the sentiment orientations that users express in reviews, sentences or on specific product features, corresponding to review(document)-level \cite{pang2002}, sentence-level \cite{sentence1,sentence2} and phrase-level \cite{contextual,optimization,lexicon2} sentiment analysis.

Review- and sentence-level sentiment analysis attempt to label a review or sentence as one of some predefined sentiment polarities, which are, typically, positive, negative and sometimes neutral \cite{survey1}. This task is referred to as Sentiment Classification, which has drawn much attention from the research community, and both supervised \cite{pang2002,mullen2004,multi-struct,compare2,hierachical,wordvec}, unsupervised \cite{turney2002,hu-kdd04,compare1,chinese1,selfsuper,chinese2} or semi-supervised \cite{semi2009,semi2010,semi2011,semi2006} methods have been investigated.

Phrase-level sentiment analysis aims to analyze the sentiment expressed by users in a finer-grained granularity. It considers the sentiment expressed on specific product features or aspects \cite{hu-kdd04}. Perhaps one of the most important tasks in phrase-level sentiment analysis is the construction of Sentiment Lexicon \cite{survey-lexicon,lexicon1,lexicon2,optimization,boost}, which is to extract feature-opinion word pairs and their corresponding sentiment polarities from these opinion rich user-generated free-texts. The construction of a high-quality sentiment lexicon would benefit various tasks, for example, personalized recommendation \cite{explain,time-profile,incorporate} and automatic review summarization \cite{hu-kdd04,survey-lexicon}.

Although some opinion words like "good" or "bad" usually express consistent sentiments in different cases, many others might have different sentiment polarities when accompanied with different feature words, which means that the sentiment lexicon is \textit{contextual} \cite{contextual}.

Various information and heuristics could be used in the process of polarity labelling of the feature-opinion pairs. For example, it is often assumed that the overall sentiment orientation of a review is aggregated from all the feature-opinion pairs in it \cite{lexicon2,optimization}. Besides, some seed opinion words that express "fixed" sentiments are usually provided, which are used to propagate the sentiment polarities of the other words \cite{hu-kdd04,optimization}. Some work takes advantage of linguistic heuristics \cite{handbook,emoticon,optimization}. For example, two feature-opinion pairs concatenated with the conjunctive "and" might have the same sentiment, while they might have opposite sentiments if connected by "but". The assumption of linguistic heuristic is further extended by sentential sentiment consistency in \cite{consistency}.

In this paper, we consider two main disadvantages of previous work. First, seldom of them combine various heuristics in a unified framework \cite{survey-lexicon,lexicon2,lexicon1,hu-kdd04,contextual}. Second, they simply use the numerical star rating as the overall sentiment polarity of the review text to supervise the process of phrase-level polarity labelling \cite{optimization,lexicon2,bayesian,peanut}.
In this work, we propose to boost phrase-level polarity labelling with review-level sentiment classification, while incorporating many of the commonly used heuristics in a unified framework.


\section{Problem Formalization}\label{sec:define}
In this section, we formalize the problems to be investigated, as well as the notations to be used in this paper.
\\\\
\textbf{Definition} (\textit{Sentiment Vector}) Suppose we are considering $r$ sentiment polarity labels $S_1,S_2\cdots S_r$, for example, \textit{positive} and \textit{negative} when $r=2$. A sentiment vector $\textbf{x}=[x_1,x_2,...,x_r]^T(x_i\ge0)$ represents the sentiment orientation of a review, sentence or feature-opinion pair. The $i$-th element $x_i$ in $\textbf{x}$ indicates the extent of sentiment on the $i$-th polarity label $S_i$. A function $s:\mathbb{R}^r\rightarrow\mathbb{R}$ is defined on sentiment vector $\textbf{x}$ such that $s(\textbf{x})$ represents the overall sentiment of the review, sentence or feature-opinion pair.\\

For example, if $S_1=positive$ and $S_2=negative$ when $r=2$, a sentiment vector $\textbf{x}=[1,0]^T$ of a review is used to indicate that the review has a sentiment orientation on \textit{positive} and no sentiment orientation on \textit{negative}. If the function $s(\textbf{x})=x_1-x_2$ is defined, then the overall sentiment orientation of the review is 1. Some previous work \cite{optimization,emoticon} enforces $x_i\le1$ as an additional constraint, however, this constraint is not necessary in our framework.
\\\\
\textbf{Definition} (\textit{Sentiment Matrix}) For a set of $m$ reviews, sentences or feature-opinion pairs $t_1,t_2\cdots t_m$, the $m\times r$ matrix $\textbf{X}=[\textbf{x}_1\textbf{x}_2\cdots\textbf{x}_m]^T$ is used to represent the sentiment orientations of them, where $\textbf{x}_i$ is the sentiment vector for $t_i$, and $s(\textbf{X})=[s(\textbf{x}_1)s(\textbf{x}_2)\cdots s(\textbf{x}_m)]^T$ is their overall sentiment orientation.
\\\\
\textbf{Definition} (\textit{Review-Level Sentiment Classification}) Given a review corpus $T$ of $m$ user reviews $t_1,t_2\cdots t_m$, a review-level sentiment classification algorithm $\mathcal{C}$ gives $\mathcal{C}(T)=\mathbf{X}_{m\times r}=[\textbf{x}_1\textbf{x}_2\cdots\textbf{x}_m]^T=[\mathcal{C}(t_1)\mathcal{C}(t_2)\cdots\mathcal{C}(t_m)]^T$, where $\mathbf{x}_i=\mathcal{C}(t_i)$ is the sentiment vector for the $i$-th user review $t_i$ given by $\mathcal{C}$.\\

Note that, in real applications, the review-level sentiment classification algorithm $\mathcal{C}$ could be both supervised or unsupervised. In this work, we consider unsupervised algorithms primarily to avoid the expensive manual labelling process and make our framework domain-independent.
\\\\
\textbf{Definition} (\textit{Sentiment Lexicon})  A sentiment lexicon constituting $n$ FO pairs $FO_1FO_2\cdots FO_n$ is defined as an $n\times r$ sentiment matrix $\textbf{X}=[\textbf{x}_1\textbf{x}_2\cdots\textbf{x}_n]^T$, where $\textbf{x}_i$ is the sentiment vector for the pair $FO_i$.\\

As has stated before, the same opinion word may express different sentiment orientations when accompanied with different feature words, which makes a sentiment lexicon \textit{contextual}. In this work, we use \textit{General Sentiment Lexicon} and \textit{Contextual Sentiment Lexicon} to indicate the two different kinds of sentiment lexicons. In the general sentiment lexicon, the sentiment vector of an FO pair is labelled according to its opinion word directly. For example, the opinion word \textit{excellent} usually expresses a positive opinion regardless of the feature word, as a result, the sentiment vector for (\textit{service},\textit{excellent}) will be labelled as $[1,0]^T$ directly. If the sentiment orientation of an opinion word is unknown, then the corresponding sentiment vector will be labelled as $[0,0]^T$. This lexicon is of high precision but low coverage. In the contextual sentiment lexicon, however, the sentiment vectors of FO pairs are labelled on considering of both the feature words and opinion words.
\\\\
\textbf{Definition} (\textit{Phrase-Level Sentiment Polarity Labelling}) In the process of contextual sentiment lexicon construction, once the feature-opinion pairs have been extracted from the review corpus, an important task is to determine the sentiment polarity of each FO pair, which is referred to as phrase-level sentiment polarity labelling. More formally, the task is to determine the sentiment matrix $\textbf{X}_{n\times r}=[\textbf{x}_1\textbf{x}_2\cdots\textbf{x}_n]^T$, which is further transformed into the overall sentiment orientations $s(\textbf{X})=[s(\textbf{x}_1)s(\textbf{x}_2)\cdots s(\textbf{x}_n)]^T$.\\

In this work, the only input of our framework is a user review corpus $T$ and a general sentiment lexicon $\textbf{X}_0$, and the expected output is a contextual sentiment lexicon $\textbf{X}$, as well as their overall sentiment orientations $s(\textbf{X})$. The general sentiment lexicon $\textbf{X}_0$ is achieved using some publicly available opinion word sets\footnote{We choose the commonly used MPQA sentiment corpus for English reviews and use HowNet for Chinese reviews. They will be formally introduced in the following part.}. These word sets contain simple and frequently used opinion words such as \textit{excellent}, \textit{good}, \textit{bad}, etc., and they are commonly viewed as basic background knowledges in natural language processing tasks.

\section{The Framework}\label{sec:framework}
In general, the framework is two-stage. In the first stage, we use an unsupervised review-level sentiment classification algorithm $\mathcal{C}$ to get the sentiment matrix $\tilde{\textbf{X}}=\mathcal{C}(T)$ for the review corpus $T$; in the second stage, we extract feature-opinion pairs from corpus $T$ and leverage the sentiment matrix $\tilde{\textbf{X}}$ as well as the general sentiment lexicon $\textbf{X}_0$ in a unified optimization framework to obtain the contextual sentiment lexicon $\textbf{X}$. After that, a pre-defined function $s(\textbf{X})$ is used to determine the overall sentiment orientation of each feature-opinion pair. We introduce the details of the framework in the following part of the section.



\subsection{Review-Level Sentiment Classification}\label{sec:unsupervised-classification}
The goal of this step is to present review-level overall sentiment polarities of user reviews by sentiment classification algorithms, with which to supervise the phrase-level sentiment polarity labelling process in the next stage.

We choose to use unsupervised review-level sentiment classification algorithms mainly because of three reasons. The first is to avoid the manual labelling process of review-level sentiment polarities, which makes our framework domain adaptable. The second is to keep our framework as general as possible to ensure its independence from specific data requirements. For example, the optimization framework in \cite{optimization} takes advantage of numerical ratings given by users in online shopping or review service websites, to supervise the process of phrase-level sentiment polarity labelling. However, such numerical ratings might not exist in specific environments, for example, in forum discussions, emails and newsgroups \cite{survey1}. Finally, the relationship between sentiment orientations of reviews and user ratings is still open \cite{wwwtutorial,survey1}. In fact, we will point out in the experiments of user rating analysis that, the numerical ratings do not necessarily indicate the sentiment orientation of textual reviews.


Classifying the sentiment orientations of user reviews into neutral or no opinion is usually ignored in most work, as has been pointed out in \cite{survey1} and \cite{wwwtutorial}, for it is extremely hard and might bring about negative effects to positive and negative sentiment classification. As a result, we choose two-class classification frameworks for review-level sentiment classification, namely, a review is classified as either positive or negative. More formally, the dimensionality of a sentiment vector is $r=2$, with the dimensionalities $S_1=positive$ and $S_2=negative$, correspondingly.

Two possible sentiment vector candidates are used in this stage. If a review is classified as $positive$ by a sentiment classification algorithm $\mathcal{C}$, then its sentient vector is assigned as $\textbf{x}=[1,0]^T$. Otherwise, the corresponding sentiment vector is set to be $\textbf{x}=[0,1]^T$. Based on the classification results, a sentiment matrix $\tilde{\textbf{X}}$ is constructed:
\begin{equation}
\tilde{\textbf{X}}=[\textbf{x}_1\textbf{x}_2\cdots\textbf{x}_m]^T
\end{equation}
where $\textbf{x}_i$ is the sentiment vector of the $i$-th review in corpus $T$. $\tilde{\textbf{X}}$ will be used as a constraint in the next stage.

%

We use the sentence orientation prediction approach in \cite{hu-kdd04} for English reviews. In this approach, a small amount (around 30) of seed opinion words are manually selected to construct the positive word set and negative word set. For example, the words such as \textit{great, fantastic, nice} and \textit{cool} are in the positive word set, and words like \textit{bad} and \textit{dull} are in the negative word set. After that, each of the two word sets are expanded by adding the synonyms of their own words and antonyms of the words from the other set, where the synonyms and antonyms are defined in WordNet. At last, the positive and negative word sets are used to aggregate the overall orientations of reviews.

We use the the automatic seed word selection scheme in \cite{chinese1} for sentiment classification of Chinese reviews. It runs on Chinese characters directly and does not require pre-segmentation. In this scheme, the positive and negative seed words are selected in an automatic framework by taking advantage of negation words, which are further used to aggregate the overall sentiment of a review.

Both of the two methods achieve pretty high sentiment classification accuracies of around $90\%$, especially on some specific domains of product reviews. For example, the precision in digital camera and mobile phone reviews could be up to $92\%$ in English corpora and $93\%$ in Chinese corpora, which are the state-of-the-art performance in sentiment classification of English and Chinese texts.

\subsection{Sentiment Lexicon Construction}
In this stage, we construct the contextual sentiment lexicon. We generate the feature-opinion pairs first, and label their polarities in a unified optimization framework.

\subsubsection{Generation of Feature-Opinion Pairs}
~\\
Each of the entries in the contextual sentiment lexicon is a feature-opinion pair. The feature word describes an aspect of a product, and the opinion word expresses the user's sentiment on the corresponding feature. In this stage, we first extract feature words from reviews, and then extract opinion words to pair with their corresponding feature words.
\\\\
\textbf{Feature words extraction}

We extract feature word candidates first, and then filter out the wrong words using a PMI-based filtering approach.

We use the Stanford Parser \cite{sparser,scparser,scdepend} to conduct Part-of-Speech tagging, morphological analysis and grammatical analysis for both English and Chinese reviews. A sentence is converted into a dependency tree after parsing, which contains both the part of speech tagging results and grammatical relationships. The following example shows the dependency tree constructed for the review sentence "\textit{Phone quality is perfect and the service is excellent.}"\\


\begin{mdframed}[innermargin=0.5cm,outermargin=0.5cm]\small
\texttt{(ROOT}

\hspace{10pt}\texttt{(S}

\hspace{20pt}\texttt{(S}

\hspace{30pt}\texttt{(NP (JJ Phone) (NN quality))}

\hspace{30pt}\texttt{(VP (VBZ is)}

\hspace{40pt}\texttt{(ADJP (JJ perfect))))}

\hspace{20pt}\texttt{(CC and)}

\hspace{20pt}\texttt{(S}

\hspace{30pt}\texttt{(NP (DT the) (NN service))}

\hspace{30pt}\texttt{(VP (VBZ is)}

\hspace{40pt}\texttt{(ADJP (JJ excellent))))}

\hspace{20pt}\texttt{(. .)))}
\end{mdframed}\vspace{10pt}

We extract the Noun Phrases (NP) and retrain those whose frequency is greater than an experimentally set threshold. These phrases are treated as feature word candidates. E.g., the phases \textit{phone quality} and \textit{the service} could be extracted.

The filtering process is then performed on the candidates in a similar way to that in \cite{extract}. We first compute the Pointwise Mutual Information (PMI) of these feature word candidates with some predefined discriminator phrases, (e.g. in the domain of cellphone the discriminator phrases are "of phone", "phone has", "phone comes with", etc). The PMI of two phrases $p_1$ and $p_2$ is defined as:
\begin{equation}\label{pmi}
\text{PMI}(p_1,p_2)=\frac{\text{Freq}(p_1,p_2)}{\text{Freq}(p_1)\cdot\text{Freq}(p_2)}
\end{equation}
where $\text{Freq}(p)$ indicates the total term frequency of a phrase $p$ in all the user reviews, and $\text{Freq}(p_1,p_2)$ is the frequency that $p_1$ and $p_2$ co-occur in a subsentence. We use subsentences instead of sentences when computing PMI because it is often the case that a sentence covers different aspects in several subsentences, as stated in \cite{optimization}. A feature word candidate is retained if its average PMI across all discriminator phrases is greater than an experimentally set threshold.
\\\\
\textbf{Opinion words extraction}

We attempt to extract opinion words from user reviews and assign them to appropriate feature words to construct feature-opinion pairs, which serve as the entries in the sentiment lexicon. For English reviews, we extract the Adjective Phrases (ADJP) that co-occur with a feature word in a subsentence as an opinion word candidate, to pair with the corresponding feature word, which forms a feature-opinion pair candidate. For example, the adjective phrase \textit{perfect} could be extracted as an opinion word, as it co-occurs with the feature word \textit{phone quality} in a subsentence. The Chinese reviews are processed in the same manner to extract opinion word candidates except that Verb Phrases (VP) are also taken into consideration besides adjective phrases.

For each of the feature-opinion pair candidates, we compute their Co-Occure Ratio (COR) in terms of the corresponding feature word. The co-occur ratio of a feature word $f$ and an opinion word $o$ is defined in the following way:
\begin{equation}\label{cor}
\text{COR}=\frac{\text{Freq}(f,o)}{\text{Freq}(f)}
\end{equation}

The notations are the same with those in equation \eqref{pmi}. Intuitionally, a high COR score means that an opinion word candidate is frequently used to modify the corresponding feature word, which indicates that they might be more likely to form a feature-opinion pair. 

We also use an experimentally set threshold of COR to filter the feature-opinion pair candidates, and the retained pairs constitute the entries in the lexicon. It is possible to employ other techniques to construct the lexicon entries, but we choose a relatively simple approach so as to focus on the next step of sentiment polarity labelling. We adopt the unified framework based on Finite State Matching Machine describe in \cite{fsm} to locate the matched feature-opinion pairs in each review sentence.

\subsubsection{Constraints on Sentiment Polarity Labelling}\label{sec:constriants}
~\\
In this step, we attempt to assign a unique sentiment polarity (\textit{positive} or \textit{negative}) to each of the feature-opinion pairs in the sentiment lexicon, using a unified convex optimization framework. The framework attempts to learn the optimal assignment of sentiment polarities by searching for the minimum solution to a loss function, where each term of the loss function is capable of representing the intuition of a specific evidence from a specific information source. Besides, the loss function is expected to be convex so that we can design a fast and simple optimization algorithm to find the unique global optimal solution to the problem.

More formally, suppose the sentiment matrix of the $n$ feature-opinion pairs extracted is represented by $\textbf{X}\in\mathbb{R}^{n\times r}$, where $r$ is the number of sentiment polarity labels used, then we attempt to learn an optimal $\textbf{X}$ and calculate the overall sentiment polarity of each pair by the function $s(\textbf{X})$. The objective function of the learning process consists of the following constrains based on different information sources.
\\\\
\textbf{Constraint on Review-level Sentiment Orientation}

Although an online user review might be either positive or negative in terms of overall sentiment polarity, it does not necessarily mean that users only discuss about positive or negative features in a single piece of review. A positive opinionated review about a product does not mean that the user has positive opinions on all aspects of the product. Likewise, a negative opinionated review does not mean that the user dislikes everything. As a result, the overall sentiment orientation of a review is the comprehensive effect of all the feature-opinion pairs contained in the review text.

Suppose we have $m$ user reviews in the review corpus $T$, then the sentiment matrix $\tilde{\textbf{X}}$ for the user reviews given by the review-level sentiment classification algorithm  is an $m\times r$ matrix, where each row of the matrix is a sentiment vector for the corresponding review.

We construct a matrix \textbf{A} with the dimension $m\times n$ to indicate the frequency of each feature-opinion pair occurring in each review. Each row of the matrix represents a review, and each column represents a feature-opinion pair. The element for review $i$ to pair $j$ is defined as:
\begin{equation}
a_{ij}=I_{ij}^{neg}\cdot\frac{\text{Freq}(i,j)}{\sum_{k}\text{Freq}(i,k)}
\end{equation}
where $\text{Freq}(i,j)$ is the frequency of feature-opinion pair $j$ in review $i$, and it would be $0$ if review $i$ does not contain pair $j$. Therefore, $\sum_{k}\text{Freq}(i,k)$ represents the total number of pairs that is contained in review $i$. The matrix $I^{neg}$ is an indication matrix that allows us to take the "negation rules" into consideration. $I_{ij}^{neg}=-1$ if the feature-opinion pair $j$ is modified by a negation word in review $i$, e.g. "no", "not", "hardly", etc. Otherwise, $I_{ij}^{neg}=1$.

According to our assumption that the overall sentiment of a text review is the comprehensive effect of all the feature opinion pairs it contains, we use a sentiment prediction function $f(\textbf{A},\textbf{X})$ to estimate the sentiment orientations of the reviews, based on the review-pairs relationship matrix $\textbf{A}$ and our contextual sentiment lexicon $\textbf{X}$. We expect to minimize the difference between our estimations and those given by the review-level sentiment classfication algorithm, which leads to the following objective function:
\begin{displaymath}\label{obj1.1}
\mathcal{R}_1=\|f(\textbf{A},\textbf{X})-\tilde{\textbf{X}}\|_F^2\setcounter{equation}{\arabic{equation}+1}\tag{\arabic{equation}.1}
\end{displaymath}

In this work, we choose a simple but frequently used sentiment prediction function, which is to predict the overall sentiment orientation of a review as the weighted average of all the feature-opinion pairs contained in it. This gives us the following objective function:
\begin{displaymath}\label{obj1.2}
\mathcal{R}_1=\|\textbf{A}\textbf{X}-\tilde{\textbf{X}}\|_F^2\tag{\arabic{equation}.2}
\end{displaymath}

As the negation words have been represented by negative weights in $\textbf{A}$, multiplying $\textbf{A}$ and $\textbf{X}$ naturally incorporates the negation rule into consideration in \eqref{obj1.2}.
\\\\
\textbf{Constraint on General Sentiment Lexicon}

As stated in the definition of sentiment lexicon, some opinion words like \textit{excellent, good} and \textit{bad} have "fixed" polarities regardless of the feature word companioned. Therefore, we construct the general sentiment lexicon $\textbf{X}_0$ by labelling the polarities of the feature-opinion pairs in $\textbf{X}$ according to publicly available sentiment corpora directly. 

We first construct a positive opinion word set and a negative opinion word set for English and Chinese reviews, respectively. The word sets for English is constructed from the MPQA opinion corpus\footnote{\texttt{http://mpqa.cs.pitt.edu/corpora/}}, which contains 2718 positive words and 4902 negative words, and the word sets for Chinese is constructed from HowNet\footnote{\texttt{http://www.keenage.com/}}, with 3730 positive words and 3116 negative words. For the $i$-th feature-opinion pair $(f,o)$, the corresponding sentiment vector $\textbf{x}_i$ in $\textbf{X}_0$ is:
\begin{equation}
\textbf{x}_i=
\begin{cases}
[1,0]^T,~if~o~is~in~positive~word~set\\
[0,1]^T,~if~o~is~in~negative~word~set\\
[0,0]^T,~otherwise
\end{cases}
\end{equation}
and finally, $\textbf{X}_0=[\textbf{x}_1\textbf{x}_2\cdots\textbf{x}_n]^T$ serves as the general sentiment lexicon.

We expect the sentiment polarities of the fixed opinion words learnt in the contextual sentiment lexicon $\textbf{X}$ to be close to those in the general sentiment lexicon $\textbf{X}_0$, which leads to the following objective function:
\begin{equation}\label{equ:obj2}
\mathcal{R}_2=\|\textbf{G}(\textbf{X}-\textbf{X}_0)\|_F^2
\end{equation}
where $\textbf{G}$ is a diagonal matrix that indicates which feature-opinion pairs in $\textbf{X}$ are "fixed" by the general sentiment lexicon $\textbf{X}_0$. Namely, $\textbf{G}_{ii}=1$ if the $i$-th feature-opinion pair has fixed sentiment, and $\textbf{G}_{ii}=0$ otherwise.
\\\\
\textbf{Constraint on Linguistic Heuristics}

An important and frequently adopted type of linguistic heuristic is the conjunctives in user reviews \cite{handbook,emoticon}. It is intuitional that feature-opinion pairs $i$ and $j$ that are frequently concatenated with "and" in the corpus might have similar sentiments, while those that are frequently connected by words like "but" tend to have opposite sentiments. For example, in the sentence "the phone quality is perfect and the sound effect is clear", if "perfect" is known to be positive, then it can be inferred that "clear" is also positive.

To formalize the intuition, we define two $n\times n$ matrices $\textbf{W}^a$ and $\textbf{W}^b$ for the "and" and "but" linguistic heuristics, respectively, where $\textbf{W}^*_{ij}\in[0,1]$ indicates our confidence that pair $i$ and $j$ have the same or opposite sentiments. A simple but frequently used choice is to set $\textbf{W}^a_{ij}=\textbf{W}^a_{ji}=1$ if pair $i$ and$j$ are concatenated by "and" for a minimal number of times in all the subsentences in the corpus, otherwise, we set $\textbf{W}^a_{ij}=\textbf{W}^a_{ji}=0$. Similarly, $\textbf{W}^b_{ij}=\textbf{W}^b_{ji}=1$ if pair $i$ and pair $j$ are linked by "but" for a minimal number of times in the corpus, and they are set to be 0 otherwise. 

To incorporate the "and" linguistic in our model, we propose to optimize the the following objective function:
\begin{equation}\label{loss_and}
\begin{aligned}
\mathcal{R}_3^a=\frac{1}{2}\sum_{i=1}^{n}\sum_{j=1}^{n}\|\textbf{X}_{i*}-\textbf{X}_{j*}\|_F^2\textbf{W}^a_{ij}=Tr(\textbf{X}^T\textbf{D}^a\textbf{X})-Tr(\textbf{X}^T\textbf{W}^a\textbf{X})
\end{aligned}
\end{equation}
where $Tr(\cdot)$ is the trace of a matrix, and $\textbf{X}_{i*}$ represents the $i$-th row of $\textbf{X}$, which is also the sentiment vector for pair $i$. $\textbf{D}^a\in\mathbb{R}^{n\times n}$ is a diagonal matrix where $\textbf{D}^a_{ii}=\sum_{j=1}^n\textbf{W}^a_{ij}$. The underlying intuition in \eqref{loss_and} is that the sentiment vectors of pairs $i$ and $j$ should be similar to each other if they are frequently linked by "and", otherwise, a penalty would be introduced to the loss function.

The formalization of "but" linguistic is similar, except that we expect the sentiment vector for pair $i$ to be close to the "opposite" of the sentiment vector for pair $j$. More intuitionally, if $\textbf{X}_{i*}$ gains a high score in its first dimension, which implies that pair $i$ tends to be positive, then $\textbf{X}_{j*}$ should also gain a high score in its second dimension, which drives pair $j$ to be negative, and vise verse.  In order to model this intuition, we introduce the following optimization term:
\begin{equation}\label{loss_but}
\begin{aligned}
\mathcal{R}_3^b=\frac{1}{2}\sum_{i=1}^{n}\sum_{j=1}^{n}\|\textbf{X}_{i*}-\textbf{X}_{j*}\textbf{E}\|_F^2\textbf{W}^b_{ij}=Tr(\textbf{X}^T\textbf{D}^b\textbf{X})-Tr(\textbf{X}^T\textbf{W}^b\textbf{XE})
\end{aligned}
\end{equation}
where $\textbf{E}=\left[\begin{smallmatrix}0&1\\1&0\end{smallmatrix}\right]$ is an anti-diagonal matrix that serves as a column permutation function to reverse the columns of $\textbf{X}$. Similarly, $\textbf{D}^b$ is a diagonal matrix where $\textbf{D}^b_{ii}=\sum_{j=1}^n\textbf{W}^b_{ij}$.

Finally, the objective function regarding both "and" and "but" linguistic heuristic is:
\begin{equation}\label{equ:obj3}
\begin{aligned}
\mathcal{R}_3&=\mathcal{R}_3^a+\mathcal{R}_3^b=Tr(\textbf{X}^T\textbf{D}^a\textbf{X})-Tr(\textbf{X}^T\textbf{W}^a\textbf{X})+Tr(\textbf{X}^T\textbf{D}^b\textbf{X})-Tr(\textbf{X}^T\textbf{W}^b\textbf{XE})\\
&=Tr(\textbf{X}^T\textbf{D}\textbf{X})-Tr(\textbf{X}^T\textbf{W}^a\textbf{X})-Tr(\textbf{X}^T\textbf{W}^b\textbf{XE})
\end{aligned}
\end{equation}
where $\textbf{D}=\textbf{D}^a+\textbf{D}^b$.
\\\\
\textbf{Constraint on Sentential Sentiment Consistency}

The use of linguistic heuristic is extended in \cite{consistency} by introducing sentential sentiment consistency (called coherency in \cite{consistency}). The fundamental assumption of sentential sentiment consistency is that the same opinion orientation (positive or negative) is usually expressed in a few consecutive sentences, which is reported to be helpful in improving the accuracy of contextual sentiment polarity labelling. 

To formalize the heuristic, a sentential similarity matrix $\textbf{W}^s\in\mathbb{R}^{n\times n}$ is introduced, which leverages the sentential distance between feature-opinion pairs in corpus to estimate their sentential similarities. For example, consider two pairs $i$ and $j$, if they co-occur in the same piece of review in the corpus, then we calculate their sentential similarity in this review, and the final similarity between $i$ and $j$ is the average of all the intra-review similarities of their co-occurrences. More formally, suppose pair $i$ and pair $j$ co-occur in the same review for $N_{ij}$ times, and the $k$-th co-occurrence happens in review $t_{i_k}$, then $\textbf{W}^s_{ij}$ and $\textbf{W}^s_{ji}$ are defined as:
\begin{equation}
\textbf{W}^s_{ij}=\textbf{W}^s_{ji}=
\begin{dcases}
0,~if~N_{ij}=0~\text{or}~\textbf{W}^a_{ij}\ne0~\text{or}~\textbf{W}^b_{ij}\ne0\\
\frac{1}{N_{ij}}\sum\limits_{k=1}^{N_{ij}}\left(1-\frac{dist(i,j)}{length(r_{i_k})}\right),~else\\
\end{dcases}
\end{equation}
where the length of a review $length(r_{i_k})$ is the number of words (punctuations excluded) in the review, and the distance of pair $i$ and $j$ in the review $dist(i,j)$ is the number of words between the two feature words of the pair. Note that we do not consider two pairs if they have been constrained by "and" ($\textbf{W}^a_{ij}\ne0$) or "but" ($\textbf{W}^b_{ij}\ne0$) linguistic heuristic. Besides, a pair might co-occur for more than one times in the same review, and we consider all the pairwise combinations in such cases.

Once the sentential similarity matrix is constructed, we incorporate sentential sentiment similarity constraint by taking into account the following objective function:
\begin{equation}\label{equ:obj4}
\begin{aligned}
\mathcal{R}_4=\frac{1}{2}\sum_{i=1}^{n}\sum_{j=1}^{n}\|\textbf{X}_{i*}-\textbf{X}_{j*}\|_F^2\textbf{W}^s_{ij}=Tr(\textbf{X}^T\textbf{D}^s\textbf{X})-Tr(\textbf{X}^T\textbf{W}^s\textbf{X})
\end{aligned}
\end{equation}
where $\textbf{D}^s$ is also a diagonal matrix, and $\textbf{D}^s_{ii}=\sum_{j=1}^n\textbf{W}^s_{ij}$.

The underlying intuition of this subjective function is that, a large penalty would be introduced if the difference of the sentiment vectors of two near pairs is significant.

\begin{algorithm}[t]
\caption{\textit{Contextual Sentiment Polarity Labelling}}
\label{alg:cspl}
\begin{algorithmic}[1]
\REQUIRE $\textbf{A}, \tilde{\textbf{X}}, \textbf{X}_0, \lambda_1, \lambda_2, \lambda_3, \lambda_4, N, \delta$
\ENSURE $\textbf{X}$
\STATE Construct matrix $\textbf{G}$ in Equation \eqref{equ:obj2}
\STATE Construct matrix $\textbf{D}, \textbf{W}^a$ and $\textbf{W}^b$ in Equation \eqref{equ:obj3}
\STATE Construct matrix $\textbf{D}^s$ and $\textbf{W}^s$ in Equation \eqref{equ:obj4}
\STATE Initialize $\textbf{X}\leftarrow\textbf{X}_0$, $\textbf{X}^{\prime}\leftarrow\textbf{X}_0$, $n\leftarrow0$
%
\REPEAT
	\STATE $n\leftarrow n+1$
	\STATE $\textbf{X}^{\prime}\leftarrow\textbf{X}$
	\FOR {each element $\textbf{X}_{ij}$ in $\textbf{X}$}
	\STATE $\textbf{X}_{ij}\leftarrow\textbf{X}_{ij}\sqrt{\frac{[\lambda_1\textbf{A}^T\tilde{\textbf{X}}+\lambda_2\textbf{G}\textbf{X}_0+\lambda_3\textbf{W}^a\textbf{X}+\lambda_3\textbf{W}^b\textbf{XE}+\lambda_4\textbf{W}^s\textbf{X}]_{ij}}{[\lambda_1\textbf{A}^T\textbf{AX}+\lambda_2\textbf{GX}+\lambda_3\textbf{DX}+\lambda_4\textbf{D}^s\textbf{X}]_{ij}}}$
	\ENDFOR
\UNTIL {$\|\textbf{X}-\textbf{X}^{\prime}\|_F^2<\delta$ \textbf{or} $n>N$}
\RETURN $\textbf{X}$
\end{algorithmic}
\end{algorithm}

\subsubsection{The Unified Model for Polarity Labelling}
~\\
With the above constraints from different information and aspects, we have the following objective function for learning the contextual sentiment lexicon $\textbf{X}$:
\begin{equation}\label{equ:obj}
\begin{aligned}
\min_{\textbf{X}\ge0}\mathcal{R}&=\lambda_1\|\textbf{A}\textbf{X}-\tilde{\textbf{X}}\|_F^2+\lambda_2\|\textbf{G}(\textbf{X}-\textbf{X}_0)\|_F^2\\
&+\lambda_3\left(Tr(\textbf{X}^T\textbf{D}\textbf{X})-Tr(\textbf{X}^T\textbf{W}^a\textbf{X})-Tr(\textbf{X}^T\textbf{W}^b\textbf{XE})\right)\\
&+\lambda_4\left(Tr(\textbf{X}^T\textbf{D}^s\textbf{X})-Tr(\textbf{X}^T\textbf{W}^s\textbf{X})\right)
\end{aligned}
\end{equation}
where $\lambda_1,\lambda_2,\lambda_3$ and $\lambda_4$ are positive weighing parameters that control the contributions of each information source in the learning process.

An important property of the objective function \eqref{equ:obj} is its convexity, which makes it possible to search for the global optimal solution $\textbf{X}^*$ to the contextual sentiment polarity labelling problem. We give the updating rule for learning $\textbf{X}^*$ directly here, as shown in \eqref{equ:update}, and the proof of the updating rule as well as its convergence is given in the appendix.
\begin{equation}\label{equ:update}
\textbf{X}_{ij}\leftarrow\textbf{X}_{ij}\sqrt{\frac{[\lambda_1\textbf{A}^T\tilde{\textbf{X}}+\lambda_2\textbf{G}\textbf{X}_0+\lambda_3\textbf{W}^a\textbf{X}+\lambda_3\textbf{W}^b\textbf{XE}+\lambda_4\textbf{W}^s\textbf{X}]_{ij}}{[\lambda_1\textbf{A}^T\textbf{AX}+\lambda_2\textbf{GX}+\lambda_3\textbf{DX}+\lambda_4\textbf{D}^s\textbf{X}]_{ij}}}
\end{equation}

The algorithm for learning the contextual sentiment lexicon is shown in Algorithm \ref{alg:cspl}. In this algorithm, we first initialize the indication matrices, Laplacian matrices and sentiment matrices through line 1 to line 4. The predefined parameter $N$ is the number of maximum iterations to conduct. The contextual sentiment lexicon $\textbf{X}$ is updated repeatedly until convergence or reaching the number of maximum iterations, where convergence means that the $\ell_2$-\textit{norm} of the difference matrix between two consecutive iterations is less than a predefined residual error $\delta$. 

\subsubsection{Overall Sentiment Polarity of the Pairs}
~\\
After the contextual sentiment lexicon $\textbf{X}$ is constructed, we use the predefined function $s(\textbf{X})=[s(\textbf{x}_1)s(\textbf{x}_2)\cdots s(\textbf{x}_n)]^T$ to determine the sentiment polarities of the feature-opinion pairs. In this work, we choose the function $s(\textbf{x}_i)=x_{i1}-x_{i2}$, where $x_{i1}$ and $x_{i2}$ are the values of positive and negative polarity labels in sentiment vector $\textbf{x}_i$, respectively. Pair $i$ is labeled as \textit{positive} if $s(\textbf{x}_i)\ge0$, and \textit{negative} if $s(\textbf{x}_i)<0$. For simplicity, we leave out the polarity label of \textit{neutral} like most of the existing work, as it is quite rare that $x_{i1}=x_{i2}$.

\section{Experiments}\label{sec:experiment}
In this section, we conduct extensive experiments to evaluate the proposed framework, and investigate the effect of different parameter settings in our framework. We will attempt to answer the following two research questions:
\begin{enumerate}
\setlength{\topsep}{0ex}
\setlength{\parskip}{0ex}
\setlength{\partopsep}{0ex}
\setlength{\itemsep}{0ex}
\setlength{\parsep}{0ex}
	\item Are the numerical star ratings always consistent with the overall sentiment orientations of textual user reviews? 
	\item How effective is our proposed framework compared with other polarity labelling methods? 
\end{enumerate}

We begin by introducing the experimental settings, and then investigate the relationship between numerical ratings and text reviews. Finally, we evaluate the proposed framework, and make some comparisons with other techniques.

\subsection{Experimental Setup}
For the experimentation on English, we use the MP3 player reviews crawled from Amazon, which is publicly available\footnote{\texttt{http://sifaka.cs.uiuc.edu/\textasciitilde wang296/Data/}}. For the experiment on Chinese language, we use the restaurant reviews crawled from DianPing.com\footnote{\texttt{http://www.dianping.com/}}, which is a famous restaurant rating website in China, and we also made this dataset publicly available\footnote{\texttt{http://tv.thuir.org/data/}}. Each of the reviews of the two datasets consists of a piece of review text and an overall numerical rating raging from 1 to 5 stars. We choose these two datasets from both English and Chinese as these two languages are of quite different types in terms of Linguistics. We want to examine whether our framework works in different language environments. Some statistical information about these two datasets is shown in Table \ref{tab:stat}.

\begin{table}[h]
\caption{Some statistics of the two datasets.}
\begin{tabular}
	{l r r r r} \hline
		&	Language	&	\#Users	&	\#Items	&	\#Reviews\\ \hline\hline
	MP3 Player	& 	English	&	26,113	&	796		&	55,740\\ \hline
	Restaurant	&	Chinese	&	11,857	&	89,462	&	510,551\\ \hline
\end{tabular}\label{tab:stat}
\end{table}

An important property of our restaurant review dataset is that, each review is accompanied with three sub-aspect ratings except for the overall rating. They are users' ratings made on the \textit{flavour}, \textit{environment} and \textit{service} of restaurants, respectively. A user is required to make ratings on all these three aspects as well as the overall experience when writing reviews on the website, which makes it possible for us to conduct much detailed user rating analysis on this dataset. The range of the sub-aspect ratings are also from 1 to 5.

\subsection{User Rating Analysis}
Previous work \cite{compare2,optimization,lara} labels a review text as positive if the corresponding overall rating is 4 or 5 stars, and negative if the overall rating is 1, 2 or 3 stars. However, the overall rating might not always be consistent with the sentient orientation of review texts. According to our observation, a substantial amount of users tend to make unaltered overall ratings although the sentiment orientation expressed in his or her review text might be quite different. Most interestingly, many users simply make 4 star ratings regardless of the review text he wrote. We analyze this phenomenon in the following part of this section.

\subsubsection{Overall and Sub-Aspect Ratings}
\begin{figure}[h]
\includegraphics[scale=0.5]{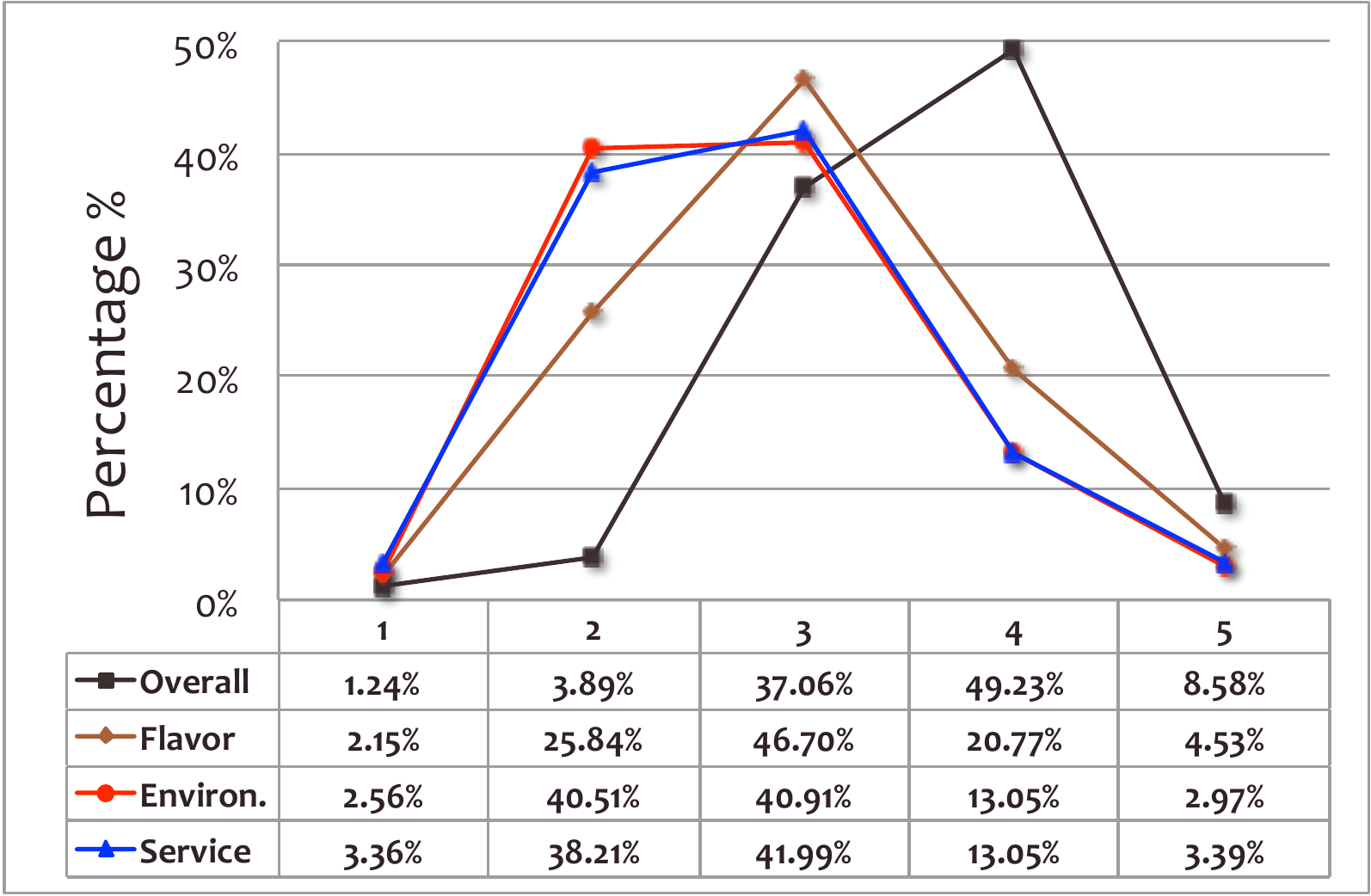}\vspace{-5pt}
\caption{The percentage of each of the five star ratings (1 star through 5 stars) against the total number of ratings, in terms of the overall rating, as well as the three kinds of sub-aspect ratings flavour, environment and service.}
\label{fig:rating-distribution}
\end{figure}

~\\
We begin by analyzing the difference in the rating distributions of the restaurant dataset.  The ratings on three sub-aspects allow us to investigate a user's "true" feelings on more specific aspects of the restaurant beyond the overall rating. For the overall rating and each of the sub-aspect ratings, we calculate the percentages that each of the 5 star ratings take in the total number of ratings, as shown in Figure \ref{fig:rating-distribution}. The x-axis represents 1 star through 5 stars, and the y-axis is the percentage of each kind of star rating.

We see that user ratings tend to center around 4 stars on overall rating, while they tend to center around 2$\sim$3 stars on the sub-aspect ratings. This implies that the overall rating might not serve as a real reflection of users' feelings, and users tend to "tell the truth" in much detailed sub-aspects. In order to examine the statistical significance, we calculate the average rating $\mu$ and coefficient of variation $c_v=\sigma/\mu$ for the overall rating and three kinds of sub-aspect ratings, where $\sigma$ is the standard deviation. Table \ref{tab:statistical-significance} shows the results. We see that users tend to give higher scores on overall rating, and the scores on overall rating are more concentrated.

\begin{table}[h]
\vspace{-10pt}
\caption{The average ratings and the coefficient of variations of the overall rating and sub-aspect ratings.}
\begin{tabular}
	{l r r r r} \hline
			&	Overall	&	Flavour	&	Environment		&	Service\\ \hline\hline
	$\mu$	& 	3.6432	&	3.1547	&	2.8934		&	2.8510\\ \hline
	$c_v$		&	0.1977	&	0.2522	&	0.2697		&	0.2816\\ \hline
\end{tabular}\label{tab:statistical-significance}
\end{table}

\begin{figure}[h]
\includegraphics[scale=0.325]{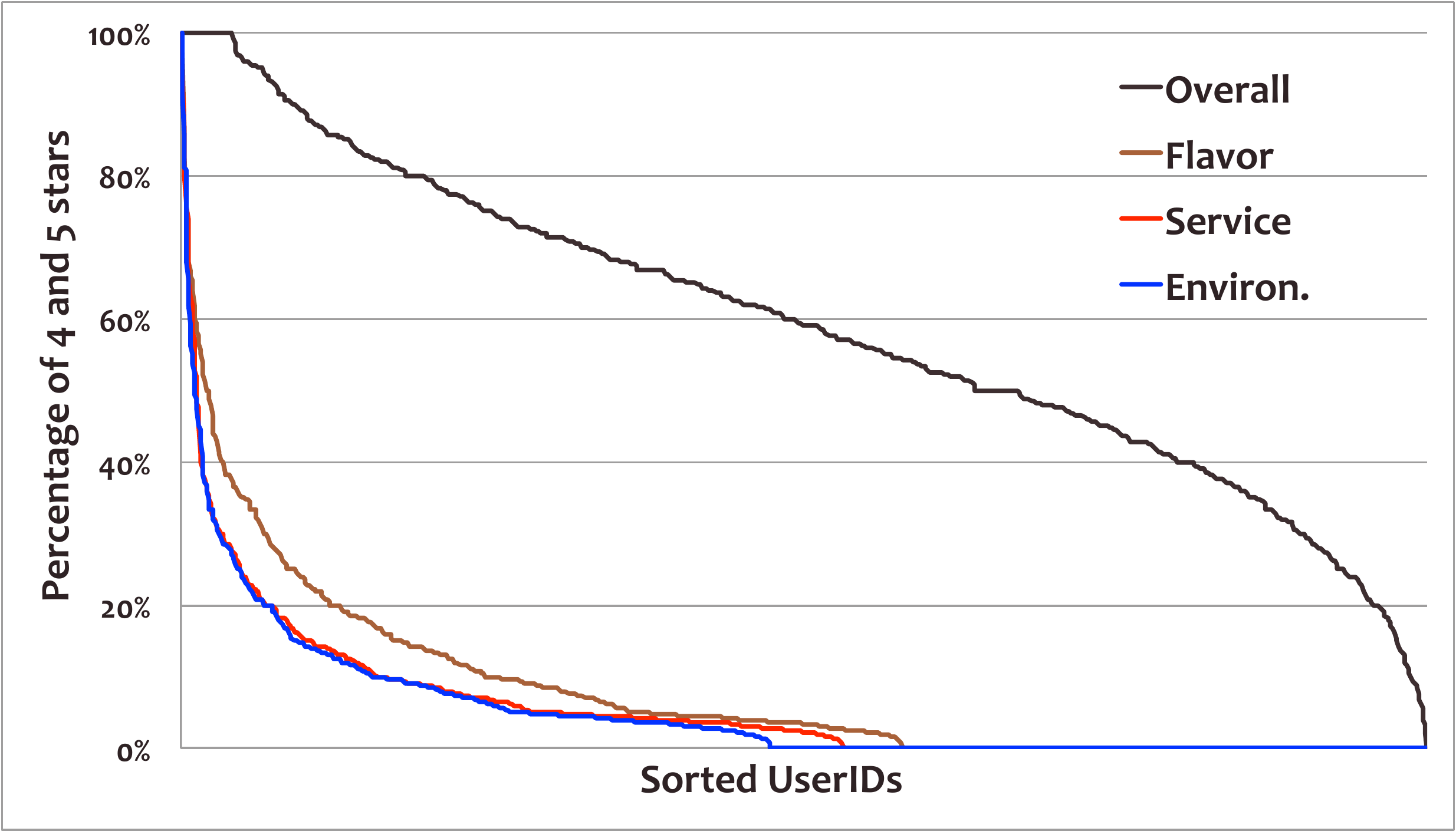}
\caption{The percentage of 4+ stars made by each of the users. The points are sorted in descending order so as to identify the fractile quantiles more easily.}
\label{fig:ratio}
\end{figure}

More intuitionally, we conduct per user analysis. For each user and each kind of rating (overall, flavour, environment and service), we calculate the percentage of 4+ stars (4 and 5 stars) that the user made. Then we sort these percentages of the users in descending order, which is shown in Figure \ref{fig:ratio}.

It is clear that user rating behaviours on overall and sub-aspect ratings are different. More than a half of the users made $50\%$ or more $4+$ ratings in terms of overall rating, while less than $5\%$ users did so on sub-aspect ratings.

This analysis partly shows that it might not be appropriate to use overall ratings as groundtruth to label the sentiment orientations of review texts, as users tend to act differently when making overall ratings and expressing their true feelings on detailed product aspects or features.

\subsubsection{Labelling Accuracy using Different Methods}
~\\
In fact, users might consider many other features except flavour, environment and service when giving the overall ratings, as a result, one may argue that the difference in distribution of ratings on different aspects is insufficient to imply that the overall rating is inappropriate in estimating the overall sentiment orientations of review texts. As a result, we evaluate the effect of sentiment orientation labelling of review texts using different kinds of numerical ratings. 

We randomly sampled 1000 reviews out of the 510,551 reviews from the restaurant review dataset to be labeled manually by 3 annotators. An annotator is asked to give a single label to a review text, which could be \textit{positive} or \textit{negative}. The final sentiment of a review is assigned by using the majority label (more than two) of the three annotators. The inter-annotator agreement is $79.76\%$, which is comparable to the reports of existing work in sentiment analysis \cite{survey2,optimization}. Finally, the annotated reviews consist of $673~(67.3\%)$ positive reviews and $327~(32.7\%)$ negative reviews.
We use four methods to estimate the overall sentiment orientation of the review texts automatically:
\begin{enumerate}
\setlength{\topsep}{0ex}
\setlength{\parskip}{0ex}
\setlength{\partopsep}{0ex}
\setlength{\itemsep}{0ex}
\setlength{\parsep}{0ex}
	\item \textbf{Overall Rating}: A review text is labeled as positive if the overall star rating is $\ge4$, and negative otherwise. The criterion of 4 stars is chosen to keep in accordance with previous work \cite{optimization,bayesian} for easier comparisons.
	\item \textbf{Normalized Overall Rating}: We use $r^\prime=r-\mu_i$ for labelling, where $r$ is the overall rating, and $\mu_i$ is the average rating of the corresponding user $i$. A review is labeled as positive if $r^\prime\ge0$ and negative if $r^\prime<0$.
	\item \textbf{Sub-Aspect Rating}: We use $\bar{r}=(r_f+r_e+r_s)/3$ for labelling, where $r_f, r_e$ and $r_s$ are the ratings on flavour, environment and service, respectively. The criterion of 4 stars is also used here.
	\item \textbf{Sentiment Classification}: We use the unsupervised review-level sentiment classification method described in section \ref{sec:unsupervised-classification} for orientation labelling.
\end{enumerate}

We use precision to evaluate the performance of each method, and the golden standard is our human annotations. The results are shown in Table \ref{tab:label-precision}, where "Pos.Review" and "Neg.Review" represent the precisions of labelling positive and negative reviews, respectively, and "Overall" is the overall performance of review-level orientation labelling.

\begin{table}[h]
\caption{The precisions of review-level sentiment orientation labelling using different methods.}
\begin{tabular}
	{l r r r r} \hline
			&	1-Overall	&	2-Normalize	&	3-Subaspect	&	4-Classify\\ \hline\hline
	Pos.Reviews 	& 	0.8321	&	0.5438	&	0.8009	&	0.9064\\ \hline
	Neg.Reviews	&	0.7248	&	0.7859	&	0.7951	&	0.8563\\ \hline
	Overall	&	0.7970	&	0.6230	&	0.7990	&	0.8900\\ \hline
\end{tabular}\label{tab:label-precision}
\end{table}

We see that the orientation labelling performance by using sub-aspect ratings (3-Subaspect) overcomes the performance of using overall ratings directly (1-Overall). However, both of their performances are worse than that given by the sentiment classification method (4-Classify).

Intuitionally, we expect to get better performance by conducting user-based normalization, as different users might have different rating scales. However, experimental results show that user-based normalization (2-Normalize) gives the worst performance. This is also because of the fact that many users make relatively high overall ratings consistently, regardless of the ratings they made on the sub-aspects. Suppose that a user made two ratings, of which one is 4 stars and the other is 5 stars. The ratings would be normalized to be -0.5 and 0.5, which leads to a negative label and a positive label. However, the 4-star review text and 5-star review text might both be positive in the golden standard.

In the following part of the experiments, we use the classification method primarily to label the review-level sentiment orientations, whose result is further used to supervise the process of contextual sentiment lexicon construction. Besides, we use the results given by overall rating labelling and sub-aspect rating labelling for performance comparison.

\subsection{Feature-Opinion Pair Extraction}
There are three experimentally set parameters in the process of feature-opinion pair extraction, which are 1) the minimum term frequency (denoted by $freq$) of Noun-Phrases to be selected as candidate feature words, 2) the minimum Pointwise Mutual  Information (denoted by $pmi$) of a feature word candidate to be retained during the filtering process, and 3) the minimum Co-Occure Ratio (denoted by $cor$) of a pair to be selected as a final feature-opinion pair.

\textbf{Pool Lexicon}: In principle, we used relatively strict parameters in order to get high quality feature-opinion pairs, which allows us to focus on the core task of phrase-level polarity labelling in this work. After careful parameter selection, we set $freq=10, pmi=0.005, cor=0.05$ on the mp3 player dataset, which leaves us with 1063 pairs, and $freq=20, pmi=0.01, cor=0.05$ on the restaurant review dataset, which gives us 1329 pairs. These pairs are presented to the three annotators for polarity labelling (positive or negative), and the final polarity of a pair is assigned according to the majority of the labels. The average agreement among annotators in this task is $81.84\%$. The pool lexicon is used for evaluating the precision of polarity labelling.

\textbf{Golden Standard Lexicon}: We then present the feature-opinion pair lists to human annotators to construct the golden standard lexicon. In this stage, each annotator is asked to select the feature-opinion pairs that describe an explicit aspect of an mp3 player or a restaurant. A pair is retained if it is selected by at least two annotators among the three, and the average agreement among annotators is $78.69\%$. The purpose of this stage is to further filter out the low quality pairs in the pool lexicon. For example, the pair (\textit{service is, good}) could be discarded as the right feature word should be \textit{service}, rather than \textit{service is}, although this pair does express a positive sentiment. The final golden standard lexicons for mp3 player dataset and restaurant review dataset consist of 695 and 857 feature-opinion pairs, respectively, and it is used for the evaluation of recall.

We use different lexicons to evaluation precision and recall to avoid the problem of evaluation bias pointed out in \cite{optimization}.

%

\subsection{Phrase-Level Polarity Labelling}
In this section, we conduct automatic sentiment polarity labelling on the feature-opinion pairs in the pool lexicon, and report the evaluation results of our method and the methods for comparison.

\subsubsection{Evaluation Measures}
~\\
We choose the frequently used measures \textit{precision, recall} and \textit{F-measure} to evaluate the performance of polarity labelling, which are defined as follows:
\begin{displaymath}
\text{precision}=\frac{N_{p\_agree}}{N_{lexicon}},~\text{recall}=\frac{N_{g\_agree}}{N_{gold}},~\text{F-measure}=\frac{2\times precision\times recall}{precision+recall}
\end{displaymath}
where $N_{lexicon}$ is the number of feature-opinion pairs in the automatically constructed sentiment lexicon, and $N_{gold}$ is the number of pairs in the golden standard lexicon (695 on the mp3 player dataset and 857 on the restaurant review dataset). $N_{p\_agree}$ and $N_{g\_agree}$ are the number of pairs consistently labeled with the pool lexicon and golden standard lexicon, respectively.

\subsubsection{Polarity Labelling Results}
~\\
We adopted the following methods for phrase-level sentiment polarity labelling in the experiment:
\begin{itemize}
\setlength{\topsep}{0ex}
\setlength{\parskip}{0ex}
\setlength{\partopsep}{0ex}
\setlength{\itemsep}{0ex}
\setlength{\parsep}{0ex}
	\item \textbf{General}: Make predictions by querying the polarity of the opinion word in general sentiment opinion word sets. Also, we use MPQA for English and HowNet for Chinese, as in section \ref{sec:constriants}. A pair is discarded if the polarity of its opinion word could not be determined.
	\item \textbf{Optimize}: The optimization approach proposed in \cite{optimization}, which reduces the problem of sentiment polarity labelling to constrained linear programming.
	\item \textbf{Overall}: Use our framework except that the review-level sentiment orientation is determined using the corresponding overall rating.
	\item \textbf{Subaspect}: Use our framework except that sentiment orientations of reviews are determined by averaging the corresponding sub-aspect ratings.
	\item \textbf{Boost}: Use our complete framework, where the sentiment classification on review text is conducted to boost phrase-level sentiment polarity labelling.
\end{itemize}

We set $\delta=0.01$ and $N=100$ in algorithm \ref{alg:cspl} to ensure convergence, and use $\lambda_1=\lambda_2=\lambda_3=\lambda_4=1$ in this set of experiment. Results on the two dataset using the above five methods are shown in Table \ref{tab:precision}, in which the bolded numbers are the best performance on the corresponding measure. We did not perform the "Subaspect" method on mp3 player reviews as the sub-aspect ratings are absent on this dataset.

%

\begin{table}[h]
\caption{Performance of sentiment polarity labelling using different methods on the MP3 player dataset (English) and restaurant review dataset (Chinese).}
\begin{tabular}
	{l r r r} \hline
			&	Precision	&	Recall	&	F-measure	\\ \hline\hline
	MP3 Player Data\\\hline\hline
	General 	& 	\textbf{0.9238}	&	0.4201	&	0.5776  \\ 
	Optimize	&	0.8269	&	0.7626	&	0.7934  \\ 
	Overall	&	0.8288	&	0.7525	&	0.7888  \\ 
	Boost		&	$^{*}0.8504$	&	\textbf{0.7683}	&	\textbf{0.8073}  \\ \hline\hline
	Restaurant Review\\\hline\hline
	General 	& 	\textbf{0.9017}	&	0.3571	&	0.5115  \\ 
	Optimize	&	0.8405	&	0.7760	&	0.8069  \\ 
	Overall	&	0.8473	&	0.7468	&	0.7938  \\ 
	Subaspect	&	0.8675	&	0.7561	&	0.8079  \\ 
	Boost		&	$^{*}0.8879$	&	\textbf{0.7818}	&	\textbf{0.8315}  \\ \hline
\end{tabular}\label{tab:precision}
\end{table}

We see that labelling the polarities by querying the general opinion word sets gives the best precision on both of the two datasets. However, the recall of this method is rather low. This implies that there are many "context dependent" opinion words which are absent from these word sets.

The "Optimize" method in \cite{optimization} and our "Overall" method are similar in that both of them leverage overall numerical ratings as the groundtruth of review-level sentiment orientations, and they make use of similar heuristics and constraints. Though the Optimize method achieves slightly better recall, their overall performance are comparable. Further more, by taking advantage of the sub-aspect ratings in the "Subaspect" method, both precision and recall are improved from "Optimize" and "Overall" methods, which implies that the detailed sub-aspect ratings could be more reliable than overall ratings.

Finally, our "Boost" method achieves the best performance in terms of recall and F-measure, on both of the two datasets. Besides, it also achieves the best precision without regard to the "General" method. This further verifies the effect of leveraging review-level sentiment classification in boosting the process of phrase-level polarity labelling.


\subsection{Parameter Analysis}
In the previous sections, we set equal weights to the different kinds of constraints for general experimental purpose. In this subsection, we attempt to study the effect of different constraints in our framework by analyzing the four main parameters $\lambda_1\sim\lambda_4$ in objective function \eqref{equ:obj}. 

We first conduct "Knock Out One Term" experiment on these parameters, to see whether all these constraints contribute to the performance of phrase-level polarity labelling. We set one of the four parameters to 0 at a time, and evaluate the F-measure. The results are shown in Table \ref{tab:drop}.

\begin{table}[h]
\caption{Evaluation results on F-measure by knocking out one constraint, where the knocked out constraint is represented by 0.}
\begin{tabular}
	{l | c c c c | c | c} \hline
		& $\lambda_1$ & $\lambda_2$ & $\lambda_3$ & $\lambda_4$ & MP3 Player & Restaurant\\ \hline\hline
	Default & 1 & 1 & 1 & 1 & 0.8073 & 0.8315 \\\hline\hline
	Knock	& 0 & 1 & 1 & 1 & 0.6783 &  0.6476	\\
	out	& 1 & 0 & 1 & 1 & 0.6332 & 0.6728 	\\
	one	& 1 & 1 & 0 & 1 & 0.7461 & 0.7352	\\
	term & 1 & 1 & 1 & 0 & 0.7756	& 0.7504	\\\hline
\end{tabular}\label{tab:drop}
\end{table}
The experimental result shows that knocking out any of the four parameters decreases the performance of polarity labelling. Besides, removing the constraint on review-Level sentiment orientation ($\lambda_1$) or the constraint on general sentiment lexicon ($\lambda_2$) decreases the performance to a great extent, which implies that these two information sources are of great importance in constructing the sentiment lexicon.

\begin{figure}[h]
\includegraphics[scale=0.42]{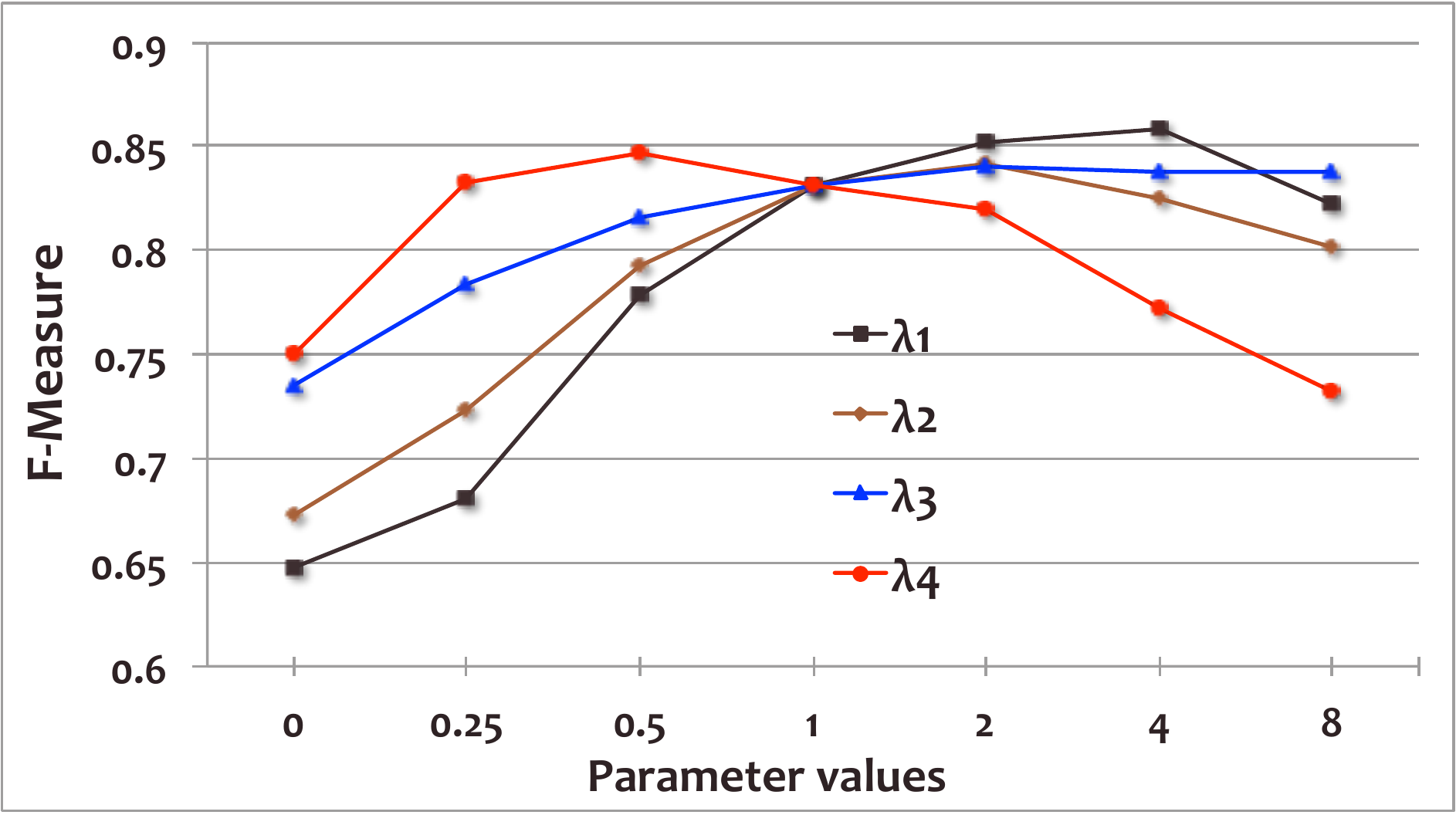}\vspace{-5pt}
\caption{Tune one of the parameters in the tuning range of $0\sim8$ with a tuning step of timing two, while fixing the remaining parameters to be 1.}
\label{fig:parameter}
\vspace{-10pt}
\end{figure}

We further investigate the effect of different constraints by fixing three parameters to 1 and weighing the remaining parameter. The experimental results on restaurant dataset are shown in Figure \ref{fig:parameter}, and the observations on mp3 player dataset are similar.

The experimental result shows that giving more weights to the constraints of review-level sentiment orientation and general sentiment lexicon could further improve the performance, which means that these two information sources might be more reliable. However, weighting the constraint on sentential sentiment consistency too much would decrease the performance, this implies that noise could be introduced by this heuristic and it is not as reliable as the linguistic heuristic of "and" and "but".

We tuned the parameters carefully to get the optimal performance. Finally, the optimal result on mp3 player dataset was achieved when using the parameters (4, 2, 1, 0.25), with an F-measure of 0.8237, and on restaurant review dataset (3, 2, 2, 0.5) is used, which gives the F-measure of 0.8584.


\section{Conclusions}\label{sec:conclusions}
Treating the numerical star rating as a sentiment indication of review text is a widely used assumption in previous phrase-level sentiment analysis algorithms \cite{optimization,bayesian,peanut}. In this paper, however, we investigated the inconsistency between the numerical ratings and textual sentiment orientations. Our observations on user rating analysis show that, users tend to make biased ratings regardless of the textual reviews they comment on a specific product. Besides, the evaluation results on labelling accuracy using different methods further verify the existence of such a bias, and indicate the effectiveness of leveraging review-level sentiment classification to recover the sentiment orientation of the reviews.

The biased assumption may hurt the performance of phrase-level sentiment polarity labelling to a large extent, because the numerical rating is usually incorporated as a kind of groundtruth to supervise the model learning process in previous work. In this paper, however, we attempt to bridge the gap between review-level and phrase-level sentiment analysis by leveraging review-level sentiment classification to boost the performance of phase-level sentiment polarity labelling. 

Specifically, we formalized the phrase-level sentiment polarity labelling problem into a simple convex optimization framework, and incorporated four kinds of heuristics to supervise the polarity labelling process. We further designed iterative optimization algorithms for model learning, where the global optimal solution is guaranteed due to the convexity of the objective function. More over, except for the four kinds of heuristics investigated in this paper, the framework is also flexible to integrate various other information sources. 

We conducted extensive experiments on two different language environments (English and Chinese) to investigate the performance of our framework, as well as its transportability across different language settings. The experimental results on both datasets show that our framework helps to improve the performance in contextual sentiment lexicon construction tasks. Besides, the experiment on parameter analysis shows that all of the four heuristics that we considered in this study contribute to the improvement in the performance of polarity labelling, which is also in accordance with previous studies.

In the future we would like to further investigate the effect of incorporating other heuristics into the framework for polarity labelling. Besides, it would be interesting to further bridge the gap between review- and phrase-level sentiment analysis by integrating the two stages into a single unified framework through, for example, deep learning techniques. Except for the sentiment polarity labelling task investigated in this work, review-level analysis could also be promising to help extract feature or opinion words in phrase-level sentiment analysis, and the joint consideration of review- and phrase-level analysis may even lead to brand new sentiment analysis tasks.



\section*{Acknowledgement}
The authors thank Prof. Xiaoyan Zhu, Yang Liu and Maosong Sun for the fruitful discussions, as well as the anonymous reviewers for the constructive suggestions. The authors also sincerely thank the annotators Yunzhi Tan, Cheng Luo and Boya Wu for their careful annotations.

\section*{Appendix}
~~~In the objective function \eqref{equ:obj}, let $\mathcal{L}=\lambda_3\textbf{D}+\lambda_4\textbf{D}^s-\lambda_3\textbf{W}^a-\lambda_4\textbf{W}^s$, let $\boldsymbol{\Lambda}$ be the Lagrange multiplier for the constraint $\textbf{X}\ge0$, and let $L(\textbf{X})$ be the Lagrange function, then we have:
\begin{equation}
\begin{aligned}
\nabla_\textbf{X}L(\textbf{X})&=2\lambda_1\textbf{A}^T\textbf{A}\textbf{X}-2\lambda_1\textbf{A}^T\tilde{\textbf{X}}+2\lambda_2\textbf{G}(\textbf{X}-\textbf{X}_0)+2\mathcal{L}\textbf{X}-2\lambda_3\textbf{W}^b\textbf{XE}-\boldsymbol{\Lambda}
\end{aligned}
\end{equation}

By setting $\nabla_\textbf{X}L(\textbf{X})=\textbf{0}$, we have:
\begin{equation}
\begin{aligned}
\boldsymbol{\Lambda}&=2\lambda_1\textbf{A}^T\textbf{A}\textbf{X}-2\lambda_1\textbf{A}^T\tilde{\textbf{X}}+2\lambda_2\textbf{G}(\textbf{X}-\textbf{X}_0)+2\mathcal{L}\textbf{X}-2\lambda_3\textbf{W}^b\textbf{XE}
\end{aligned}
\end{equation}

According to the Karush-Kuhn-Tucker (KKT) complementary condition \cite{convex} on the non-negativity constraint on $\textbf{X}$, we have $\boldsymbol{\Lambda}_{ij}\cdot\textbf{X}_{ij}=0$, namely:
\begin{equation}\label{equ:kkt1}
\begin{aligned}
[\lambda_1\textbf{A}^T\textbf{A}\textbf{X}&-\lambda_1\textbf{A}^T\tilde{\textbf{X}}+\lambda_2\textbf{G}(\textbf{X}-\textbf{X}_0)+\mathcal{L}\textbf{X}-\lambda_3\textbf{W}^b\textbf{XE}]_{ij}\cdot\textbf{X}_{ij}=0
\end{aligned}
\end{equation}

Equation \eqref{equ:kkt1} can be further transformed into:
\begin{equation}
\begin{aligned}
[-(\lambda_1\textbf{A}^T\tilde{\textbf{X}}&+\lambda_2\textbf{G}\textbf{X}_0+\lambda_3\textbf{W}^a\textbf{X}+\lambda_3\textbf{W}^b\textbf{XE}+\lambda_4\textbf{W}^s\textbf{X})\\
&+(\lambda_1\textbf{A}^T\textbf{AX}+\lambda_2\textbf{GX}+\lambda_3\textbf{DX}+\lambda_4\textbf{D}^s\textbf{X})]_{ij}\cdot\textbf{X}_{ij}=0
\end{aligned}
\end{equation}

which leads to the updating rule of $\textbf{X}$ as follows:
\begin{equation}
\textbf{X}_{ij}\leftarrow\textbf{X}_{ij}\sqrt{\frac{[\lambda_1\textbf{A}^T\tilde{\textbf{X}}+\lambda_2\textbf{G}\textbf{X}_0+\lambda_3\textbf{W}^a\textbf{X}+\lambda_3\textbf{W}^b\textbf{XE}+\lambda_4\textbf{W}^s\textbf{X}]_{ij}}{[\lambda_1\textbf{A}^T\textbf{AX}+\lambda_2\textbf{GX}+\lambda_3\textbf{DX}+\lambda_4\textbf{D}^s\textbf{X}]_{ij}}}
\end{equation}

The correctness and convergence of the updating rule can be proved using the standard auxiliary function approach presented in \cite{nmf}.

\bibliographystyle{fullname}
\bibliography{COLI-manual1}

\begin{thebibliography}{}

\bibitem[\protect\citename{Bickerstaffe and Zukerman}2010]{hierachical}
Bickerstaffe, Adrian and Ingrid Zukerman.
\newblock 2010.
\newblock {A Hierarchical Classifier Applied to Multi-way Sentiment Detection}.
\newblock {\em Proceedings of the 21st International Conference on
  Computational Linguistics (Coling)}, pages 62--70.

\bibitem[\protect\citename{Boyd and Vandenberghe}2004]{convex}
Boyd, S. and L.~Vandenberghe.
\newblock 2004.
\newblock {Convex Optimization}.
\newblock {\em Cambridge University Press}.

\bibitem[\protect\citename{Chang \bgroup et al.\egroup }2009]{scdepend}
Chang, Pichuan, Huihsin Tseng, Dan Jurafsky, and Christopher~D. Manning.
\newblock 2009.
\newblock {Discriminative Reordering with Chinese Grammatical Relations
  Features}.
\newblock {\em Proceedings of the Third Workshop on Syntax and Structure in
  Statistical Translation (SSST)}, pages 51--59.

\bibitem[\protect\citename{Cui, Mittal, and Datar}2006]{compare2}
Cui, Hang, Vibhu Mittal, and Mayur Datar.
\newblock 2006.
\newblock {Comparative Experiments on Sentiment Classification for Online
  Product Reviews}.
\newblock {\em Proceedings of the 21st national conference on Artificial
  intelligence (AAAI)}, 2:1265--1270.

\bibitem[\protect\citename{Dasgupta and Ng}2009]{semi2009}
Dasgupta, Sajib and Vincent Ng.
\newblock 2009.
\newblock {Mine the Easy, Classify the Hard: A Semi-Supervised Approach to
  Automatic Sentiment Classification}.
\newblock {\em Proceedings of the 47th Annual Meeting of the Association for
  Computational Linguistics (ACL)}, 2:701--709.

\bibitem[\protect\citename{Dave, Lawrence, and Pennock}2003]{peanut}
Dave, Kushal, Steve Lawrence, and David~M. Pennock.
\newblock 2003.
\newblock {Mining the Peanut Gallery: Opinion Extraction and Semantic
  Classification of Product Reviews}.
\newblock {\em WWW}, pages 519--528.

\bibitem[\protect\citename{Ding, Liu, and Yu}2008]{lexicon2}
Ding, Xiaowen, Bing Liu, and Philip~S. Yu.
\newblock 2008.
\newblock {A Holistic Lexicon-Based Approach to Opinion Mining}.
\newblock {\em Proceedings of the 2008 International Conference on Web Search
  and Data Mining (WSDM)}, pages 231--239.

\bibitem[\protect\citename{Goldberg and Zhu}2006]{semi2006}
Goldberg, Andrew~B. and Xiaojin Zhu.
\newblock 2006.
\newblock {Seeing stars when there aren't many stars: Graph-based
  Semi-supervised Learning for Sentiment Categorization}.
\newblock {\em Proceedings of the First Workshop on Graph Based Methods for
  Natural Language Processing}, pages 45--52.

\bibitem[\protect\citename{Hu and Liu}2004]{hu-kdd04}
Hu, Minqing and Bing Liu.
\newblock 2004.
\newblock {Mining and Summarizing Customer Reviews}.
\newblock {\em Proceedings of the 10th ACM SIGKDD international conference on
  Knowledge discovery and data mining (KDD)}, pages 168--177.

\bibitem[\protect\citename{Hu \bgroup et al.\egroup }2013]{emoticon}
Hu, Xia, Jiliang Tang, Huiji Gao, and Huan Liu.
\newblock 2013.
\newblock {Unsupervised Sentiment Analysis with Emotional Signals}.
\newblock {\em WWW}, pages 607--617.

\bibitem[\protect\citename{Jansen \bgroup et al.\egroup }2009]{wom}
Jansen, Bernard~J., Mimi Zhang, Kate Sobel, and Abdur Chowdury.
\newblock 2009.
\newblock {Micro-blogging as Online Word of Mouth Branding}.
\newblock {\em Proceedings of the 2009 International Conference on Human
  Factors in Computing Systems (CHI)}, pages 3859--3864.

\bibitem[\protect\citename{Kanayama and Nasukawa}2006]{consistency}
Kanayama, Hiroshi and Tetsuya Nasukawa.
\newblock 2006.
\newblock {Fully Automatic Lexicon Expansion for Domain-oriented Sentiment
  Analysis}.
\newblock {\em Proceedings of the 2006 Conference on Empirical Methods in
  Natural Language Processing (EMNLP)}, pages 355--363.

\bibitem[\protect\citename{Lee and Seung}2001]{nmf}
Lee, Daniel~D. and H.~Sebastian Seung.
\newblock 2001.
\newblock {Algorithms for Non-negative Matrix Factorization}.
\newblock {\em Proceedings of the Neural Information Processing Systems
  (NIPS)}, pages 556--562.

\bibitem[\protect\citename{Levy and Manning}2003]{scparser}
Levy, Roger and Christopher~D. Manning.
\newblock 2003.
\newblock {Is it harder to parse Chinese, or the Chinese Treebank?}
\newblock {\em Proceedings of the 43rd Annual Meeting of the Association for
  Computational Linguistics (ACL)}, pages 439--446.

\bibitem[\protect\citename{Li \bgroup et al.\egroup }2011]{semi2011}
Li, Shoushan, Zhongqing Wang, Guodong Zhou, and Sophia Yat~Mei Lee.
\newblock 2011.
\newblock {Semi-Supervised Learning for Imbalanced Sentiment Classification}.
\newblock {\em Proceedings of the 22nd International Joint Conference on
  Artificial Intelligence (IJCAI)}, 3:1826--1831.

\bibitem[\protect\citename{Lin, He, and Everson}2010]{compare1}
Lin, Chenghua, Yulan He, and Richard Everson.
\newblock 2010.
\newblock {A Comparative Study of Bayesian Models for Unsupervised Sentiment
  Detection}.
\newblock {\em Proceedings of the 14th Conference on Computational Natural
  Language Learning (CoNLL)}, pages 144--152.

\bibitem[\protect\citename{Liu}2010]{handbook}
Liu, Bing.
\newblock 2010.
\newblock {Sentiment Analysis and Subjectivity}.
\newblock {\em Handbook of Natural Language Processing, Chapman and Hall/CRC, 2
  edition}.

\bibitem[\protect\citename{Liu, Hu, and Cheng}2005]{lexicon1}
Liu, Bing, Minqing Hu, and Junsheng Cheng.
\newblock 2005.
\newblock {Opinion Observer: Analyzing and Comparing Opinions on the Web}.
\newblock {\em WWW}, pages 342--351.

\bibitem[\protect\citename{Liu and Zhang}2012]{survey1}
Liu, Bing and Lei Zhang.
\newblock 2012.
\newblock {A Survey of Opinion Mining and Sentiment Analysis}.
\newblock {\em Mining Text Data}, pages 415--463.

\bibitem[\protect\citename{Liu, Seneff, and Zue}2010]{os1}
Liu, Jingjing, Stephanie Seneff, and Victor Zue.
\newblock 2010.
\newblock {Dialogue-Oriented Review Summary Generation for Spoken Dialogue
  Recommendation Systems}.
\newblock {\em Proceedings of the 2010 Annual Conference of the North American
  Chapter of the Association for Computational Linguistics (NAACL)}, pages
  64--72.

\bibitem[\protect\citename{Lu \bgroup et al.\egroup }2011]{optimization}
Lu, Yue, Malu Castellanos, Umeshwar Dayal, and ChengXiang Zhai.
\newblock 2011.
\newblock {Automatic Construction of a Context-Aware Sentiment Lexicon: An
  Optimization Approach}.
\newblock {\em WWW}, pages 347--356.

\bibitem[\protect\citename{Lu, Zhai, and Sundaresan}2009]{bayesian}
Lu, Yue, ChengXiang Zhai, and Neel Sundaresan.
\newblock 2009.
\newblock {Rated Aspect Summarization of Short Comments}.
\newblock {\em WWW}, pages 131--140.

\bibitem[\protect\citename{M.~Marneffe}2006]{sparser}
M.~Marneffe, B.~Maccartney, C.~Manning.
\newblock 2006.
\newblock {Generating Typed Dependency Parses from Phrase Structure Parses}.
\newblock {\em Proceedings of the 5th International Conference on Language
  Resources and Evaluation (LREC)}, pages 449--454.

\bibitem[\protect\citename{Maas \bgroup et al.\egroup }2011]{wordvec}
Maas, Andrew~L., Raymond~E. Daly, Peter~T. Pham, Dan Huang, Andrew~Y. Ng, and
  Christopher Potts.
\newblock 2011.
\newblock {Learning Word Vectors for Sentiment Analysis}.
\newblock {\em Proceedings of the 51st Annual Meeting of the Association for
  Computational Linguistics (ACL)}, pages 142--150.

\bibitem[\protect\citename{Moghaddam and Ester}2013]{wwwtutorial}
Moghaddam, Samaneh and Martin Ester.
\newblock 2013.
\newblock {Opinion Mining in Online Reviews: Recent Trends}.
\newblock {\em Tutorial on the 22th international conference on World Wide
  Web}.

\bibitem[\protect\citename{Mullen and Collier}2004]{mullen2004}
Mullen, Tony and Nigel Collier.
\newblock 2004.
\newblock {Sentiment Analysis using Support Vector Machines with Diverse
  Information Sources}.
\newblock {\em Proceedings of the 2004 Conference on Empirical Methods in
  Natural Language Processing (EMNLP)}, pages 412--418.

\bibitem[\protect\citename{Nakagawa, Inui, and Kurohashi}2010]{sentence2}
Nakagawa, Tetsuji, Kentaro Inui, and Sadao Kurohashi.
\newblock 2010.
\newblock {Dependency Tree-based Sentiment Classification using CRFs with
  Hidden Variables}.
\newblock {\em Proceedings of the 2010 Annual Conference of the North American
  Chapter of the Association for Computational Linguistics (NAACL)}, pages
  786--794.

\bibitem[\protect\citename{Orimaye, Alhashmi, and Siew}2013]{om}
Orimaye, Sylvester~O., Saadat~M. Alhashmi, and Eu~Gene Siew.
\newblock 2013.
\newblock {Performance and Trends in Recent Opinion Retrieval Techniques}.
\newblock {\em The Knowledge Engineering Review}, pages 1--30.

\bibitem[\protect\citename{Pang and Lee}2008]{survey2}
Pang, Bo and Lillian Lee.
\newblock 2008.
\newblock {Opinion Mining and Sentiment Analysis}.
\newblock {\em Foundations and Trends in Information Retrieval}, 2(1-2):1--135.

\bibitem[\protect\citename{Pang, Lee, and Vaithyanathan}2002]{pang2002}
Pang, Bo, Lillian Lee, and Shivakumar Vaithyanathan.
\newblock 2002.
\newblock {Thumbs up? Sentiment Classification using Machine Learning
  Techniques}.
\newblock {\em Proceedings of the 2002 Conference on Empirical Methods in
  Natural Language Processing (EMNLP)}, pages 79--86.

\bibitem[\protect\citename{Popescu and Etzioni}2005]{extract}
Popescu, Ana~Maria and Oren Etzioni.
\newblock 2005.
\newblock {Extracting Product Features and Opinions from Reviews}.
\newblock {\em Proceedings of the 2005 Conference on Empirical Methods in
  Natural Language Processing (EMNLP)}, pages 339--346.

\bibitem[\protect\citename{Qiu \bgroup et al.\egroup }2009]{selfsuper}
Qiu, Likun, Weishi Zhang, Changjian Hu, and Kai Zhao.
\newblock 2009.
\newblock {SELC: A Self-Supervised Model for Sentiment Classification}.
\newblock {\em Proceedings of the 18th ACM Conference on Information and
  Knowledge Management (CIKM)}, pages 929--936.

\bibitem[\protect\citename{Taboada \bgroup et al.\egroup }2011]{survey-lexicon}
Taboada, Maite, Julian Brooke, Milan Tofiloski, Kimberly Voll, and Manfred
  Stede.
\newblock 2011.
\newblock {Lexicon-Based Methods for Sentiment Analysis}.
\newblock {\em Computational Linguastics}, 37(2):267--307.

\bibitem[\protect\citename{Tan \bgroup et al.\egroup }2013]{fsm}
Tan, Yunzhi, Yongfeng Zhang, Min Zhang, Yiqun Liu, and Shaoping Ma.
\newblock 2013.
\newblock {A Unified Framework for Emotional Elements Extraction Based on
  Finite State Matching Machine}.
\newblock {\em Natural Language Processing and Chinese Computing (NLP\&CC)},
  pages 60--71.

\bibitem[\protect\citename{Turney}2002]{turney2002}
Turney, Peter~D.
\newblock 2002.
\newblock {Thumbs Up or Thumbs Down? Sentiment Orientation Applied to
  Unsupervised Classification of Reviews}.
\newblock {\em Proceedings of the 42nd Annual Meeting of the Association for
  Computational Linguistics (ACL)}, pages 417--424.

\bibitem[\protect\citename{Wang, Lu, and Zhai}2010]{lara}
Wang, Hongning, Yue Lu, and Chengxiang Zhai.
\newblock 2010.
\newblock {Latent Aspect Rating Analysis on Review Text Data: A Rating
  Regression Approach}.
\newblock {\em KDD}, pages 783--792.

\bibitem[\protect\citename{Wiebe, Wilson, and Cardie}2005]{sentence1}
Wiebe, Janyce, Theresa Wilson, and Claire Cardie.
\newblock 2005.
\newblock {Annotating Expressions of Opinions and Emotions in Language}.
\newblock {\em Language Resources and Evaluation}, 39:165--210.

\bibitem[\protect\citename{Wilson, Wiebe, and Hoffmann}2005]{contextual}
Wilson, Theresa, Janyce Wiebe, and Paul Hoffmann.
\newblock 2005.
\newblock {Recognizing Contextual Polarity in Phrase-Level Sentiment Analysis}.
\newblock {\em Proceedings of the 2005 Conference on Empirical Methods in
  Natural Language Processing (EMNLP)}, pages 347--354.

\bibitem[\protect\citename{Yessenalina, Yue, and Cardie}2010]{multi-struct}
Yessenalina, Ainur, Yisong Yue, and Claire Cardie.
\newblock 2010.
\newblock {Multi-level Structured Models for Document-level Sentiment
  Classification}.
\newblock {\em Proceedings of the 2010 Conference on Empirical Methods in
  Natural Language Processing (EMNLP)}, pages 1046--1056.

\bibitem[\protect\citename{Zagibalov and Carroll}2008a]{chinese1}
Zagibalov, Taras and John Carroll.
\newblock 2008a.
\newblock {Automatic Seed Word Selection for Unsupervised Sentiment
  Classification of Chinese Text}.
\newblock {\em Proceedings of the 19st International Conference on
  Computational Linguistics (Coling)}, pages 1073--1080.

\bibitem[\protect\citename{Zagibalov and Carroll}2008b]{chinese2}
Zagibalov, Taras and John Carroll.
\newblock 2008b.
\newblock {Unsupervised Classification of Sentiment and Objectivity in Chinese
  Text}.
\newblock {\em IJCNLP}, pages 304--311.

\bibitem[\protect\citename{Zhang}2015]{incorporate}
Zhang, Yongfeng.
\newblock 2015.
\newblock {Incorporating Phrase-level Sentiment Analysis on Textual Reviews for
  Personalized Recommendation}.
\newblock {\em Proceedings of the 8th ACM international conference on Web
  Search and Data Mining (WSDM)}.

\bibitem[\protect\citename{Zhang \bgroup et al.\egroup }2014a]{explain}
Zhang, Yongfeng, Guokun Lai, Min Zhang, Yi~Zhang, Yiqun Liu, and Shaoping Ma.
\newblock 2014a.
\newblock {Explicit Factor Models for Explainable Recommendation based on
  Phrase-level Sentiment Analysis}.
\newblock {\em Proceedings of the 37th international ACM SIGIR conference on
  Research \& development in information retrieval (SIGIR)}, pages 83--92.

\bibitem[\protect\citename{Zhang \bgroup et al.\egroup }2014b]{boost}
Zhang, Yongfeng, Haochen Zhang, Min Zhang, Yiqun Liu, and Shaoping Ma.
\newblock 2014b.
\newblock {Do Users Rate or Review? Boost Phrase-level Sentiment Labeling with
  Review-level Sentiment Classification}.
\newblock {\em Proceedings of the 37th international ACM SIGIR conference on
  Research \& development in information retrieval (SIGIR)}, pages 1027--1030.

\bibitem[\protect\citename{Zhang \bgroup et al.\egroup }2015]{time-profile}
Zhang, Yongfeng, Min Zhang, Yi~Zhang, Guokun Lai, Yiqun Liu, and Shaoping Ma.
\newblock 2015.
\newblock {Daily-Aware Personalized Recommendation based on Feature-Level Time
  Series Analysis}.
\newblock {\em Proceedings of the 24nd international conference on World Wide
  Web (WWW)}.

\bibitem[\protect\citename{Zhou, Chen, and Wang}2010]{semi2010}
Zhou, Shusen, Qingcai Chen, and Xiaolong Wang.
\newblock 2010.
\newblock {Active Deep Networks for Semi-Supervised Sentiment Classification}.
\newblock {\em Proceedings of the 21st International Conference on
  Computational Linguistics (Coling)}, pages 1515--1523.

\end{thebibliography}

\end{document}